\documentclass{article}

\usepackage{PRIMEarxiv}

\usepackage[utf8]{inputenc} 
\usepackage[T1]{fontenc}    
\usepackage{hyperref}       
\usepackage{url}            
\usepackage{booktabs}       
\usepackage{amsfonts}       
\usepackage{nicefrac}       
\usepackage{microtype}      
\usepackage{lipsum}
\usepackage{fancyhdr}       
\usepackage{graphicx}       
\graphicspath{{media/}}     
\usepackage{graphicx}
\usepackage{amsmath}
\usepackage{amssymb}
\usepackage{booktabs}
\usepackage{url}

\usepackage{amsmath,amsfonts,bm}

\def\eqref#1{equation~\ref{#1}}

\def\1{\bm{1}}

\DeclareMathAlphabet{\mathsfit}{\encodingdefault}{\sfdefault}{m}{sl}
\SetMathAlphabet{\mathsfit}{bold}{\encodingdefault}{\sfdefault}{bx}{n}

\usepackage{subcaption}

\title{Deployment of a Robust and Explainable Mortality Prediction Model: The COVID-19 Pandemic and Beyond
}

\author{
  Jacob R. Epifano, Stephen Glass, Ravi P. Ramachandran \\
  Department of Electrical and Computer Engineering \\
  Rowan Univsercity \\
  Glassboro NJ, USA\\
  \texttt{epifanoj0@rowan.edu sglass520@gmail.com, ravi@rowan.edu} \\
  \And
  Sharad Patel \\
  Department of Critical Care Medicine\\
  Cooper University Hospital\\
  Camden NJ, USA\\
  \texttt{patel-sharad@cooperhealth.edu}
   \And
  Aaron J. Masino \\
  Department of Biostatistics, Epidemiology, Informatics \\
  University of Pennsylvania Perelman School of Medicine \\
  Philadelphia PA, USA\\
  \texttt{aaron.masino@pennmedicine.upenn.edu} \\
   \And
  Ghulam Rasool \\
  Department of Machine Learning \\
  Moffitt Cancer Center \\
  Tampa FL, USA\\
  \texttt{ghulam.rasool@moffitt.org} \\
}

\begin{document}
\maketitle

\begin{abstract}
This study investigated the performance, explainability, and robustness of deployed artificial intelligence (AI) models in predicting mortality during the COVID-19 pandemic and beyond. The first study of its kind, we found that Bayesian Neural Networks (BNNs) and intelligent training techniques allowed our models to maintain performance amidst significant data shifts. Our results emphasize the importance of developing robust AI models capable of matching or surpassing clinician predictions, even under challenging conditions. Our exploration of model explainability revealed that stochastic models generate more diverse and personalized explanations thereby highlighting the need for AI models that provide detailed and individualized insights in real-world clinical settings. Furthermore, we underscored the importance of quantifying uncertainty in AI models which enables clinicians to make better-informed decisions based on reliable predictions. Our study advocates for prioritizing implementation science in AI research for healthcare and ensuring that AI solutions are practical, beneficial, and sustainable in real-world clinical environments. By addressing unique challenges and complexities in healthcare settings, researchers can develop AI models that effectively improve clinical practice and patient outcomes. 
\end{abstract}


\section{Introduction}
Artificial Intelligence (AI) and Machine Learning (ML) have started to gain traction in medical domains \cite{mesko2020short}. However, due to the high-stakes nature of medicine, AI/ML models rarely make it to real-world deployments \cite{vayena2018machine}. While a few models have successfully transitioned to deployment \cite{smith2013ability, masino2019machine, elish2018stakes}, user feedback is often negative \cite{bedoya2019minimal, guidi2015clinician}. Factors such as high false alarm rates and uninterpretable outputs \cite{dewan2020performance, ginestra2019clinician} demand several considerations for AI/ML to flourish in high-risk environments like healthcare.

The challenges of AI/ML in healthcare are well-recognized in the field \cite{tonekaboni2019clinicians}. Consequently, substantial efforts have been made to distill missing elements into a few key requirements. Many emphasize the importance of interpretability and explainability \cite{jimenez2020drug, cutillo2020machine, combi2022manifesto}. 
Interpretability is concerned with understanding the inner workings of a model and is frequently associated with transparent design \cite{tonekaboni2019clinicians}. 
Explainability concentrates on the comprehension of how and why a model decision was arrived at. However, with the advent of deep learning, ML practitioners are increasingly trading the transparent design of linear models, whose decisions are easily understood (e.g., logistic regression, decision trees), for the better performance of black-box models (e.g., Deep Neural Networks (DNNs)) \cite{london2019artificial}. In a black-box model, the inputs and outputs are observable. However, the hidden nature of the algorithmic operations gives no idea of how the outputs are arrived at. As a result, a significant body of work has been dedicated to explaining the predictions of black-box DNN models. In particular, several notable methods exist for explaining the predictions of DNNs due to their popularity. Nonetheless, there is a growing concern regarding how accurate an explainability method is (known as faithfulness \cite{nielsen2023evalattai}) as these methods rely on creating local interpretable models for each prediction \cite{rudin2019stop}. 

Measuring a model's confidence or its uncertainty is equally important in medical AI \cite{jimenez2020drug, tonekaboni2019clinicians}. 
DNNs have been shown to be typically overconfident in their predictions \cite{nguyen2015deep, guo2017calibration}, i.e., high softmax output does not equate to high prediction confidence \cite{ahmed2022failure}. This becomes crucial when clinicians must take action based on a prediction, as their subsequent actions may vary depending on their perception of the situation's severity \cite{umscheid2015development}. Overconfident models lead to high false-alarm rates, which undermine overall trust in the AI model \cite{embi2012evaluating}.

\begin{figure*}[tb]
     \centering
     \begin{subfigure}[b]{0.23\textwidth}
         \centering
         \includegraphics[width=\textwidth]{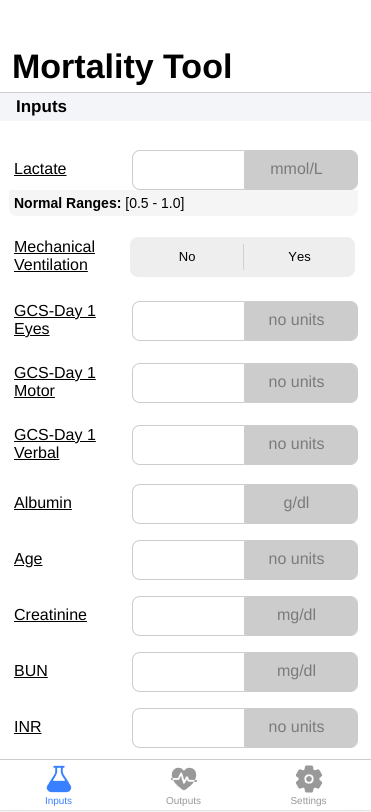}
         \caption{}
         \label{subfig:app_input}
     \end{subfigure}
     \hfill
     \begin{subfigure}[b]{0.23\textwidth}
         \centering
         \includegraphics[width=\textwidth]{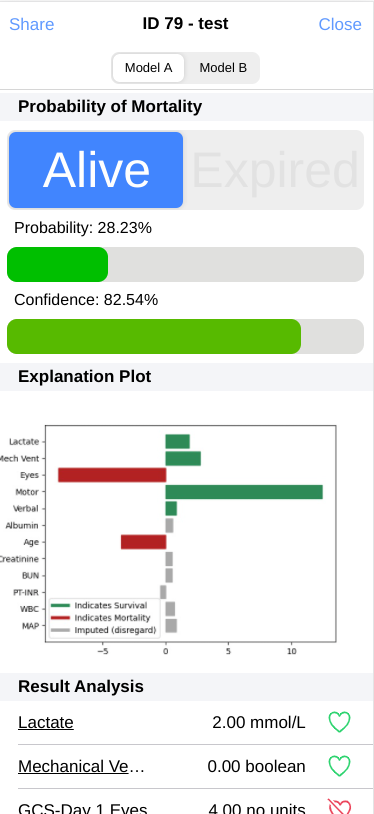}
         \caption{}
         \label{subfig:app_output}
     \end{subfigure}
     \hfill
     \begin{subfigure}[b]{0.23\textwidth}
         \centering
         \includegraphics[width=\textwidth]{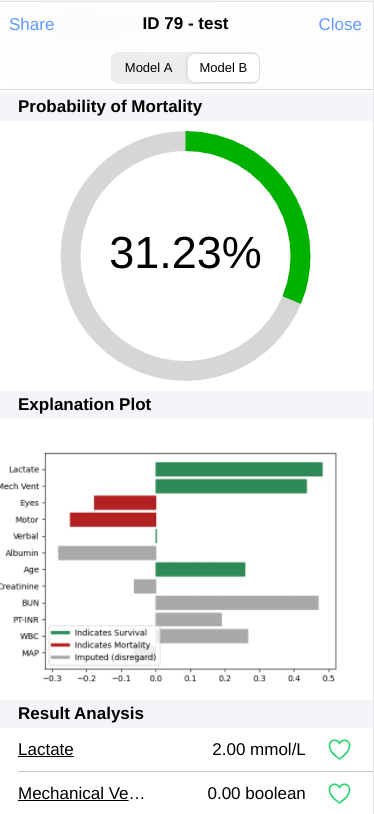}
         \caption{}
         \label{subfig:app_output2}
     \end{subfigure}
     \hfill
     \begin{subfigure}[b]{0.23\textwidth}
         \centering
         \includegraphics[width=\textwidth]{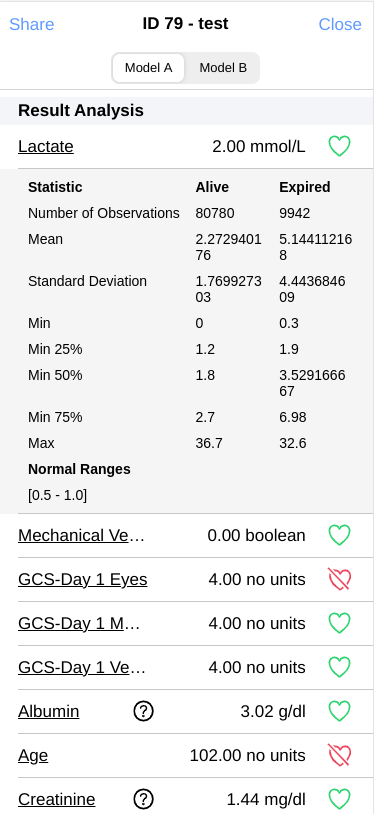}
         \caption{}
         \label{subfig:app_stats}
     \end{subfigure}
     \caption{Images of web-app user interface. \textbf{\ref{subfig:app_input}:} input screen. All variables can be selected and the healthy range for each variable will appear in a drop-down menu. This feature was requested by our users. \textbf{\ref{subfig:app_output}}: output of model A (VDP). This output pane includes the probability of mortality as well as the confidence score derived from the variance output of the stochastic network. \textbf{\ref{subfig:app_output2}:} output of model B (deterministic/traditional neural net). Below the output probabilities for each model, an explanation plot derived from influence functions is provided. For each input feature, the influence function provides an indication of whether or not the signal contributed towards one class or the other (indicated by the sign) and the magnitude determines how much each signal contributed to the classification. \textbf{\ref{subfig:app_stats}:} dataset statistics. For our users who want a more fine grained understanding of the classification decision, the output screen contains a drop down menu where it shows statistics from the training set. This may provide the user with a better understanding of how the model came to its prediction.}
    \label{fig:app_pics}
\end{figure*}

Lastly, some studies highlight dataset-shift as a crucial factor to consider when designing AI/ML models for healthcare \cite{kelly2019key}. Dataset-shift occurs most commonly when the model is deployed in a data or environment that is different from the one used during training. A robust model is insensitive to dataset-shift.
Dataset-shift is often overlooked during the design and development process or is left for manual review and update. It is naive to assume that the independent and identically distribution (i.i.d.) assumption of many AI/ML algorithms will hold in a field like healthcare, where different patient populations can exhibit vastly different distributions \cite{nestor2018rethinking}. 

Algorithmic bias is another related and critical concern in healthcare.
Algorithmic bias exists when a repeatable error leads to an unfair outcome.
It has been demonstrated that mortality prediction models' performance can vary depending on patient ethnicity \cite{chen2018my}. 
There are common issues in the mitigation of dataset-shift and algorithmic bias.
The two collectively address the full spectrum of failure of black-box AI models.

Our contributions include:
\begin{enumerate}
    \item Real world deployment of a first-day mortality prediction model designed with the help of critical care clinicians.
    \item Usage of a noise-robust Bayesian Neural Network to quantify prediction uncertainty via covariance propagation.
    \item Implementation and deployment of Influence Functions \cite{koh2017understanding} to provide instance-level explanations, a trait highly desired by clinicians \cite{tonekaboni2019clinicians}.
    \item A post-hoc analysis of model performance during significant data-shift induced by the COVID-19 
    pandemic thereby highlighting the robust behavior of our models.
\end{enumerate}

\begin{figure*}[tb]
     \centering
     \begin{subfigure}[b]{0.31\textwidth}
         \centering
         \includegraphics[width=\textwidth]{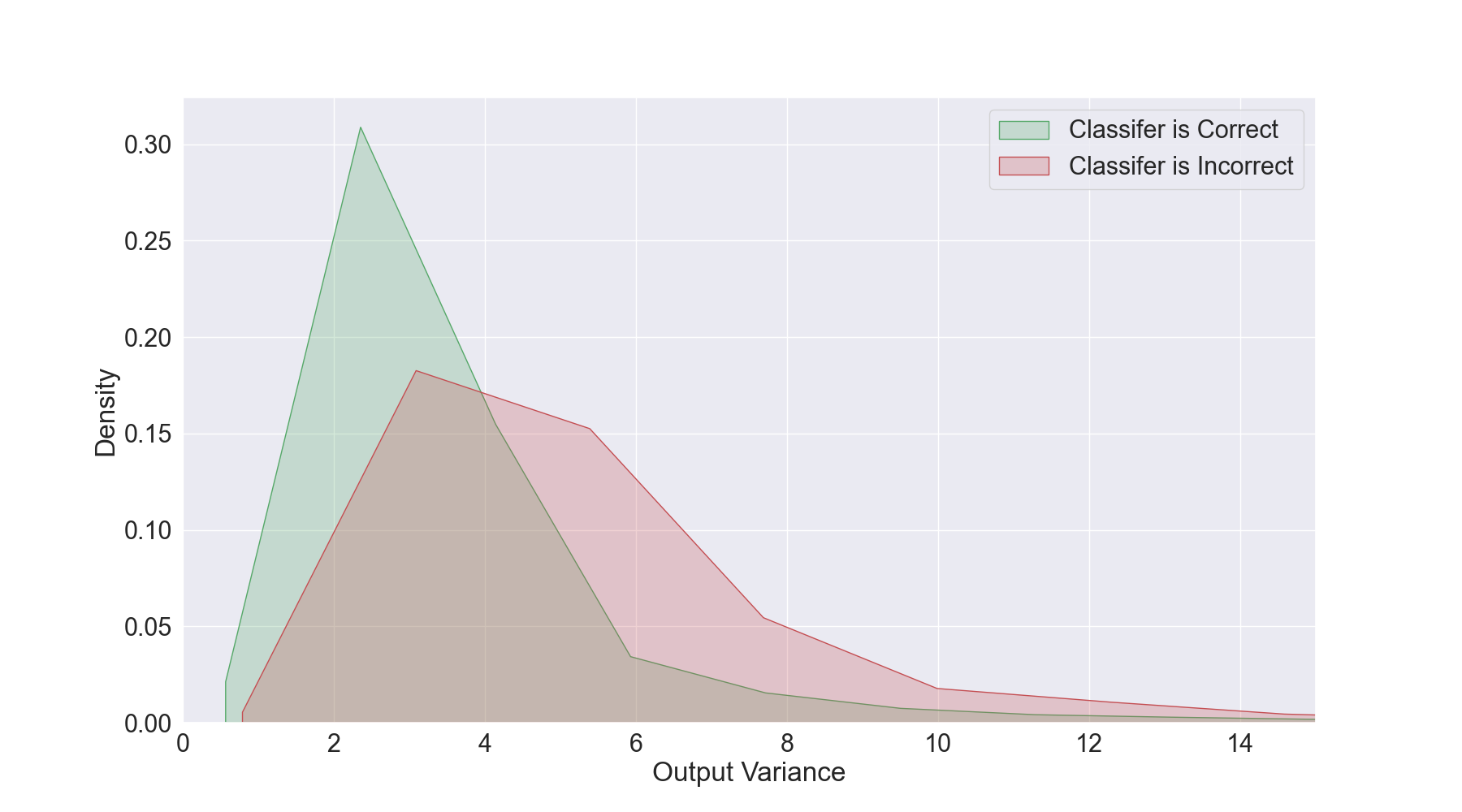}
         \caption{}
         \label{subfig:train_sigma}
     \end{subfigure}
     \hfill
     \begin{subfigure}[b]{0.31\textwidth}
         \centering
         \includegraphics[width=\textwidth]{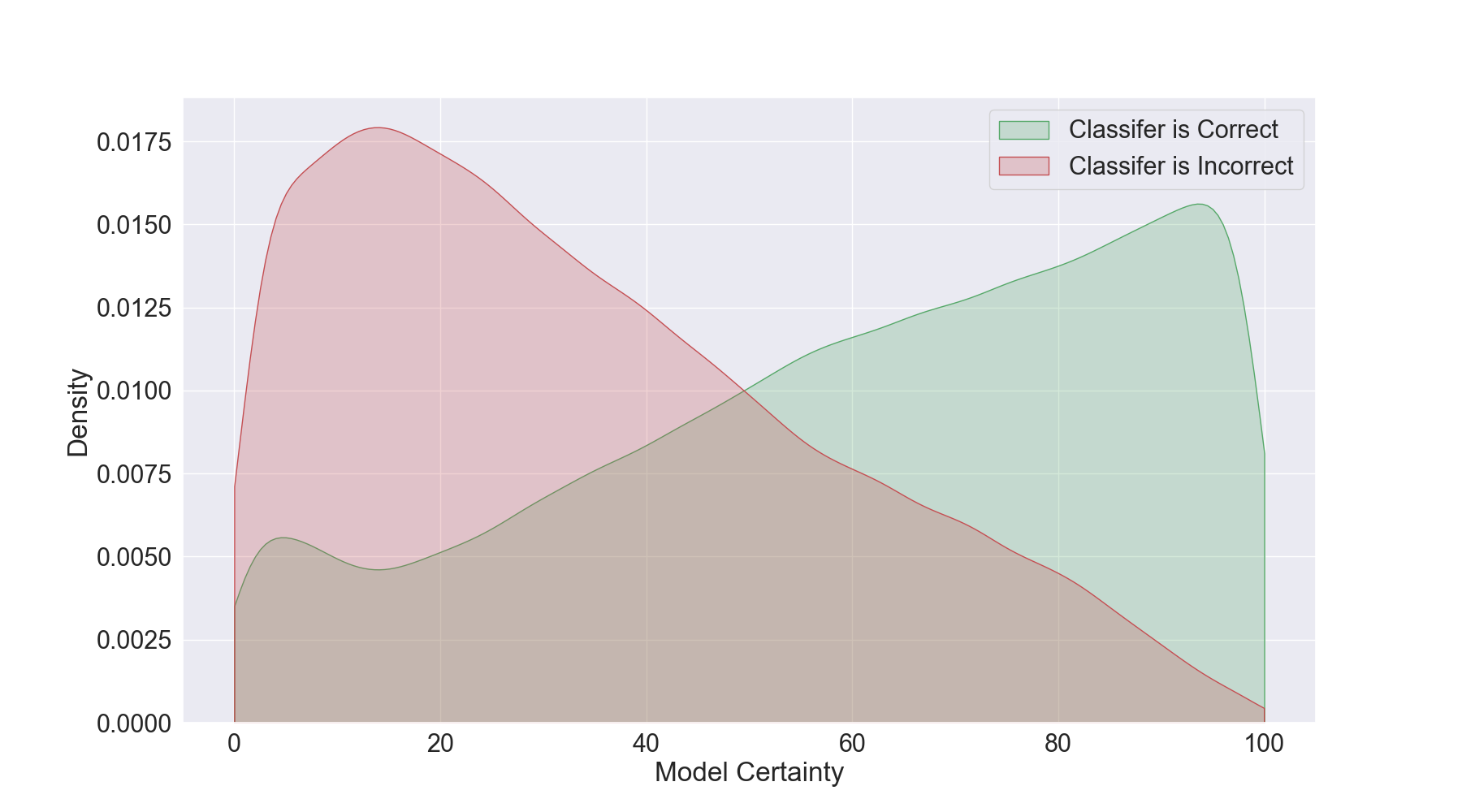}
         \caption{}
         \label{subfig:train_uncertainty}
     \end{subfigure}
     \hfill
     \begin{subfigure}[b]{0.31\textwidth}
         \centering
         \includegraphics[width=\textwidth]{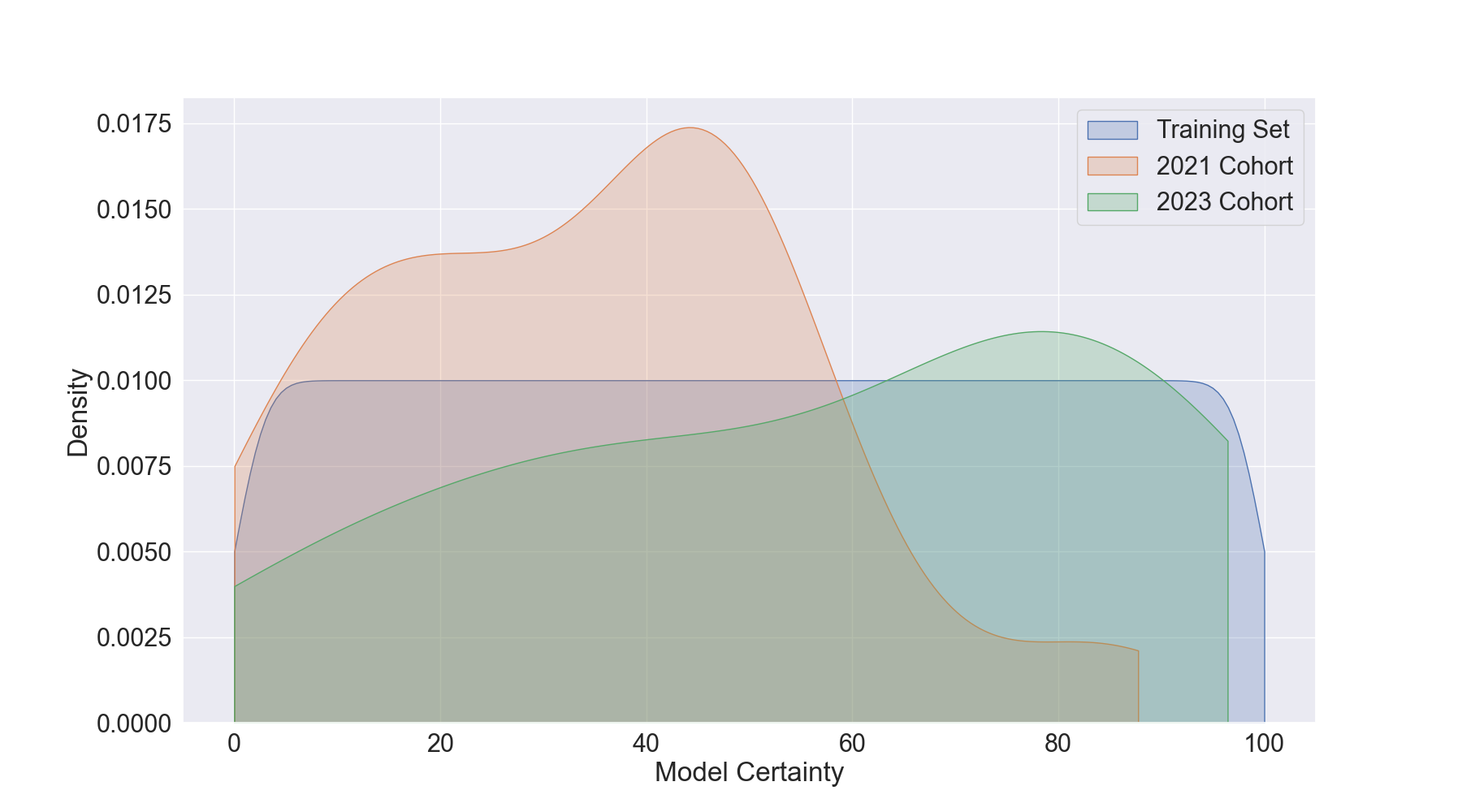}
         \caption{}
         \label{subfig:cohort_uncertainty}
     \end{subfigure}
    \caption{\textbf{\ref{subfig:train_sigma}:} Output variance and \textbf{\ref{subfig:train_uncertainty}:} Uncertainty scores computed across training set by estimating the Cummulative Density Function (CDF) of the joint distribution. On average, the variance is lower (model is more confident) when the prediction is correct. This indicates that the Bayesian model is well calibrated to the dataset. \textbf{\ref{subfig:cohort_uncertainty}:} Uncertainty scores broken down by cohort. COVID-19 afflicted cohort (2021) has lower certainty than unafflicted cohort (2023). The 2023 cohort seemingly follows the training distribution.}
    \label{fig:sigma_dist}
\end{figure*}

\section{Related Work}

\subsection{Explainable AI}
The performance of healthcare AI/ML models has started to become a secondary priority to the interpretability of the model. A tool that only outputs a probability score for the presence of a disease is not one that would likely be endorsed by clinical staff. Several approaches have become mainstream and have made their way into clinical workflows \cite{bedoya2019minimal, guidi2015clinician}. The most popular class of explainable methods are feature attribution methods or methods which try to attribute the prediction to one or several input features. Influence functions \cite{koh2017understanding} have been used in mortality prediction \cite{epifano2020towards}. Also, SHapley Additive exPlanations (SHAP) values \cite{lundberg2017unified} were used in drug discovery \cite{rodriguez2019interpretation}  and first-day mortality prediction \cite{safaei2022catboost, epifano2020towards}. The power behind feature attribution methods is that they can often be used to estimate the global feature importance as well as give local feature importance for a given instance. These two elements were called out specifically by clinician surveys as being highly desirable for a given clinical model \cite{tonekaboni2019clinicians}. 

\subsection{Uncertainty Estimation}
When discussing uncertainty, it is important to distinguish between the two sub-types. The first is  aleatoric uncertainty, which refers to the uncertainty of the data or the environment. The second is epistemic uncertainty, which refers to the uncertainty of the parameters of the model \cite{hullermeier2021aleatoric}. In a clinical setting, it is becoming highly desirable to include an estimate of the uncertainty of the prediction \cite{jimenez2020drug, tonekaboni2019clinicians}. This is typically done via ensemble approaches like Deep ensemble averaging \cite{lakshminarayanan2017simple} or  probabilistic approaches like Bayesian Neural Nets (BNNs) \cite{blundell2015weight, dera2019extended}. Baking model uncertainty into models comes with benefits. Techniques like stochastic weight averaging \cite{izmailov2018averaging} and Varitational Density Propagation (VDP) \cite{dera2019extended} boast better generalization and robustness to noise and artifacts. To our knowledge, models equipped with uncertainty estimation have yet to make it to production settings \cite{kompa2021second}. 

\subsection{Dataset-shift and algorithmic bias}
One obvious example of dataset-shift in healthcare came with the advent of COVID-19. A particular issue is mortality prediction in the Intensive Care Unit (ICU). During several periods of the pandemic, ICU mortality rates spiked \cite{auld2022trends}. Several biomarkers were found to be better predictors of mortality in COVID-19 patients than standard AI/ML models \cite{yan2020interpretable}. This same model was replicated and tried in the Netherlands and it was found that the mortality prediction accuracy was much lower \cite{quanjel2021replication}. This case study shows that even when a tool is tailor made to serve a specific subset of the population, it may only work well in a specific region. When developing these models, it can become easy to fool oneself about the performance of a model. Several strategies should be employed once the model is deployed. These include periodical testing using data from the deployment, model fine-tuning, and full model retraining \cite{davis2019nonparametric}.


\section{Methods}

\subsection{Dataset and Training Scheme}
To train our model, we aggregated mortality data from two publicly available datasets: MIMIC-III \cite{johnson2016mimic} and the eICU \cite{pollard2018eicu} databases. From these databases, we selected data such as demographic information (age, gender), indicator variables (patient on mechanical ventilation, metastatic cancer, etc.) and lab values (blood lactate, glucose, creatinine, etc.) that are given within 24 hours of patient admittance. This results in nearly 200,000 instances and 200 features with a 92\%-8\% class imbalance where patient mortality is the positive class. In light of limited sample size, to reduce the likelihood of model overfitting, we adopted an aggressive feature selection procedure. First, pair-wise correlations were computed and all features with Pearson correlation coefficient greater than 0.9 were dropped from the dataset. Next, any features with greater than 50\% missingness were removed from the dataset. We then computed the mutual information between the remaining features and the mortality outcome and selected the top 20. Out of the top 20, a group of clinicians chose 12 based on clinical opinion. We opted to not take a purely data science feature selection technique since some of the top features chosen did not give clinicians any actionable insights. Additionally by taking user opinion into account during the model selection stage, we hoped to foster stronger understanding between model and user prior to deployment. Our final set of features is as follows: Blood lactate, mechanical ventilation (yes/no), all Glasgow Coma Scale parameters (eyes, motor, verbal), albumin, age, creatinine, Prothrombin Time (PT-INR), White Blood cell Count (WBC), Blood Urea Nitrogen (BUN), and Mean Arterial Pressure (MAP). 

Model training was performed in a similar manner to \cite{epifano2020towards} where first-day mortality prediction was investigated. First-day prediction of mortality is a useful indicator for clinicians in order to direct care, although trained models may be used any day after admittance. Based on input of our clinician collaborators, missing data was assumed to be missing at random. We used multiple imputation with chained equations \cite{van2011mice} as the imputation method. Prior to training, the data was standardized. The learning algorithm chosen here is a shallow neural network. To optimize parameters and capture performance metrics, we used a 10-fold cross validation scheme to split our data into train and test sets. We used Weights and Biases \cite{wandb} to optimize the hyper parameters of our neural network. After discussing with our clinician collaborators, we chose to optimize for positive likelihood ratio (LR+ = sensitivity / (100 – specificity)) as it is a metric commonly used for diagnostic tests and therefore readily understood by our users \cite{deeks2004diagnostic}. Given the large class imbalance, various options to rectify the imbalance were given to the hyperparameter optimizer: Synthetic Minority Oversampling (SMOTE) \cite{chawla2002smote, JMLR:v18:16-365}, majority undersampling \cite{JMLR:v18:16-365}, and loss function weighting (penalizing minority class predictions). After the hyperparameters are selected, we trained the models using both splits. Once deployed, we did not perform any retraining or online leaning.

\subsection{Variational Density Propagation (VDP)}
In addition to our traditional or deterministic neural net, we also trained a Bayesian Neural Network (BNN) in the same manner. We decided to include this model in order to investigate whether or not using robust models will aid performance in non-stationary environments. We will refer to this as the stochastic neural net. We used Variational Density Propagation (VDP) \cite{dera2019extended, dera2020bayes, dera2021premium}, which is based on Tensor Normal Distributions \cite{manceur2013maximum} and variational inference. This method has several advantages over other BNNs such as Bayes By Backprop \cite{blundell2015weight} due to the propagation of the first two moments (mean and covariance) through the network. These advantage are: noise robustness, model averaging, and Uncertainty Quantification (UQ). By including this model we can satisfy the UQ constraint outlined by the clinician survey \cite{tonekaboni2019clinicians} without relying on an external UQ metrics that may not be faithful to the original model. A full derivation of the VDP framework is provided in \ref{sec:vdp_dev}.

\begin{table}[tp]
\centering
\begin{tabular}{|c|c|c|}
\hline
\textbf{Model}    & \textbf{ROC AUC} & \textbf{Number of Input Features} \\ \hline
\cite{le1993new}           & 0.83             & 15                                \\ \hline
\cite{zimmerman2006acute}        & 0.84             & 37                                \\ \hline
\cite{epifano2020towards}  & 0.85             & 20                                \\ \hline
\cite{safaei2022catboost} & 0.86$\pm$(0.02)  & 10                                \\ \hline
\cite{epifanofeature} & 0.90$\pm$(0.00)  & 20                                \\ \hline
Determistic (Ours)       & 0.89$\pm$(0.00)  & 12                                \\ \hline
Stochastic (Ours)        & 0.88$\pm$(0.00)  & 12                                \\ \hline
\end{tabular}
\caption{10-fold cross validated training results for the traditional (deterministic) and Bayesian \cite{dera2019extended} (stochastic) neural nets compared to several baseline models. All models were designed to predicted first-day mortality and were trained and evaluated on the eICU dataset. The mean score across the 10 runs is shown with the standard error.}
\label{tab:training_res}
\end{table}

To quantify the uncertainty of our prediction, we transform the output of the network. The output of the model contains the mean and covariance. The variance (diagonal elements of the covariance) contain the information about the uncertainty of the prediction. Since this value ranges from $[0, \infty)$, it can be difficult to understand the uncertainty of any one given prediction. To demonstrate how the variance can be used to estimate uncertainty, we have have plotted the distribution of variance outputs (from our training set) by whether or not the classifier correctly classified the instance. We should expect lower variance output when the classifier is correct (less noise in the prediction) than when it is incorrect. Figure \ref{subfig:train_sigma} shows these distributions. In order to make this metric understandable to our users, we estimate the CDF of the empirical variance distribution of our training set, $1-\textrm{CDF}(\sigma^2)$. This metric will be bounded in the range $[0, 1]$, where 0 indicates low certainty and 1 indicates high certainty (Figure \ref{subfig:train_uncertainty}).

\subsection{Explanations and Interpretability}
To satisfy the interpretability constraints requested by clinicians \cite{tonekaboni2019clinicians}, we provide instance-level explanations via influence functions for each prediction. This method was previously thought to be fragile \cite{basu2020influence}, but has been recently shown to work for small networks \cite{epifano2023revisiting}. LIME \cite{ribeiro2016should} and SHAP \cite{lundberg2017unified} are the most commonly used methods to explain black-box models, however they rely on creating a surrogate model and don't provide guarantees that their explanations will be faithful to the original model \cite{slack2020fooling, lakkaraju2020fool, fernandez2020explaining}. Another major weakness of other methods like SHAP, is that they fail when multicollinearity exists in the dataset used to train the model. We solve this problem with our aggressive feature selection scheme which firsts removes highly correlated features. Feature attribution-based explanations are still useful after our feature selection scheme as our final feature set consists of signals that are orthogonal along the treatment space. Determining the most impactful signals can lead to actionable insights for our users. 


Consider a standard classification problem where a label $y$ is predicted for each feature vector $x$. Let $z_i = (x_i, y_i)$, where $i = 1, 2, ..., N$, for $N$ instances in the dataset. It is assumed that we have a trained model where $\theta$ represents the trained network parameters. Our loss function can be written as $L(z, \theta) = \sum_{i=1}^N L(z_i, \theta)$. Our optimal model parameters are the set of parameters that minimize the loss: $\hat{\theta} = \textrm{arg}\min_{\theta \in \Theta}\sum_{i=1}^N L(z_i, \theta)$
\cite{koh2017understanding}. Recall, from Koh and Liang \cite{koh2017understanding}, influence functions estimate the parameter change, $\hat{\theta}_{-z} - \hat{\theta}$ due to removing a training instance $z$. The estimated parameter change is given in Equation \ref{eq:up_params}, where $H_{\hat{\theta}} = \frac{1}{n}\sum_{i=1}^n \nabla^2_\theta L(z_i, \hat{\theta})$ is the Hessian \cite{koh2017understanding}. Using the chain rule, we can estimate the change in loss on a test instance, $z_\textrm{test}$ using Equation \ref{eq:up_loss} \cite{koh2017understanding}. Finally, by perturbing the input features of the training instance to be removed, we can estimate the effect of the perturbation on the loss of the test instance, given by Equation \ref{eq:pert_loss} \cite{koh2017understanding}.

\begin{equation}
    \mathcal{I}_{\textrm{up,params}}(z)= -H^{-1}_{\hat{\theta}}\nabla_\theta L(z,\hat{\theta}),
    \label{eq:up_params}
\end{equation}

\begin{equation}
    \mathcal{I}_{\textrm{up,loss}}(z, z_{\textrm{test}}) = -\nabla_\theta L(z_{\textrm{test}}, \hat{\theta})^\top H_{\hat{\theta}}^{-1}\nabla_\theta L(z, \hat{\theta})
    \label{eq:up_loss}
\end{equation}

\begin{equation}
    \mathcal{I}_{\text{pert,loss}}(z,z_{\text{test}})^\top=  -\nabla_\theta L(z_{\text{test}},\theta)^\top H_\theta^{-1}\nabla_x \nabla_\theta L(z,\theta).
    \label{eq:pert_loss}
\end{equation}

By aggregating Equation \ref{eq:pert_loss} over the training set, we can find the average effect that perturbing a single input feature has on loss of the test instance. This local feature importance (FI) score is given in Equation \ref{eq:local} \cite{epifano2020towards}. Explanations generated by influence functions are ideal for this application as the loss difference provides a magnitude and direction indicating feature importance and sentiment, giving insight into which class (in the binary case) the model believes the feature is attributed to.

\begin{equation}
    FI_{\text{local}} = \frac{1}{N}\sum_{i=1}^N I_{\text{pert,loss}}(z_i,z_{\text{test}}).
    \label{eq:local}
\end{equation}

In addition to the explanations, we have taken steps to make the model more interpretable using the app User Interface (UI). As 
suggested by clinicians during development, normative ranges for each feature are provided via drop down on the input screen as shown in Figure \ref{subfig:app_input}. We have also computed some first order statistics for each feature in our dataset in an effort to foster understanding between the model and users. These statistics are provided via drop down on the output screen as shown in Figure \ref{subfig:app_stats}.

\subsection{App Design for Android/Apple}
The design of the app can be split into three components: front-end client, API server, and ML background worker. The parts are deployed in a monolithic architecture on a Google Compute Engine virtual machine (VM) with a graphical processing unit (GPU). Ionic Framework and React library using TypeScript language was used to develop the front-end UI as a Progressive Web Application. The application can be accessed by a publicly accessible domain on a web browser and optionally installed on the device. With communication to the API server, the UI allows users to create accounts, input data using the interface, and view or update previous records with retrospective analysis. Images of the app's UI can be found in Figure \ref{fig:app_pics}.

\section{Results}
\subsection{Model Selection}
Upon the conclusion of model selection, the optimal parameters for each model were as follows. For the deterministic network, the model had three hidden layers with widths 197, 198, and 112 nodes, respectively. The loss function was weighted to account for the large class imbalance with a positive weight of 14.80. The model was trained for 127 epochs of full-batch gradient descent using stochastic gradient descent (SGD) with parameters: learning rate $\gamma=0.03104$, weight-decay $\lambda=0.0104$, and Nesterov momentum $\mu=0.4204$. 

The stochastic model also had three hidden layers with widths 31, 93, and 94 respectively. The class imbalance was handled by majority undersampling. The model was trained for 18 epochs with a batch-size of 1000 using mini-batch SGD with parameters: learning rate $\gamma=0.0022$, weight-decay $\lambda=0.0064$, and Nesterov momentum $\mu=0.7589$. We performed a t-test on the test statistics of the two models and no significant difference was found ($p>0.01$).

Table \ref{tab:training_res} compares our cross-validated model performance against other first-day mortality prediction models. The metric typically used to compare mortality prediction models is the Receiver Operating Characteristic Area Under the Curve (ROC AUC) \cite{zimmerman2006acute}. Both of our models achieve a significantly higher ROC AUC than readily available (typically via online calculator) scores (APACHE IV \cite{zimmerman2006acute}, SAPS \cite{le1993new}), as well as a recently published Neural Network model \cite{epifano2020towards} and ensemble model \cite{safaei2022catboost} ($p< 0.01$ for all pair-wise t-tests). Our models fall behind a recent Elastic-Net model \cite{epifanofeature}, however more input features were required to achieve this performance. Additional statistics and comparisons of our two models on the union of the eICU and MIMIC-III databases are included in Appendix \ref{sec:metrics}. 

\subsection{Deployment and data collection}
Model deployment commenced in January 2021, during which data from two distinct cohorts were gathered. The first cohort consisted of 59 subjects, with data collected between January 2021 and November 2021 (COVID-19 surge). The second cohort included 25 subjects, and their data were obtained from January 2023 to May 2023 (Post COVID-19). In both cohorts combined, a total of 43 subjects had their mortality outcomes documented (21 from 2021 and 22 from 2023). This was accompanied by predictions from a critical care clinician prior to viewing model predictions and true outcomes. We chose to collect data in this way in order to study the effects that dataset drift may have on our models due to the COVID-19 pandemic. 

\begin{table}[tp]
\centering
\resizebox{\textwidth}{!}{%
\begin{tabular}{|c|c|cc|cc|}
\hline
Classifier   & Clinician & \multicolumn{2}{c|}{Deterministic Network}             & \multicolumn{2}{c|}{Stochastic Network}                \\ \hline
Datatset     & Accuracy  & \multicolumn{1}{c|}{ROC AUC}         & Accuracy        & \multicolumn{1}{c|}{ROC AUC}         & Accuracy        \\ \hline
eICU $\cup$ MIMIC-III & -         & \multicolumn{1}{c|}{0.87$\pm$(0.00)} & 0.77$\pm$(0.00) & \multicolumn{1}{c|}{0.87$\pm$(0.00)} & 0.77$\pm$(0.00) \\ \hline
2021 Cohort  & 0.95      & \multicolumn{1}{c|}{\textbf{0.99}}   & \textbf{0.90}   & \multicolumn{1}{c|}{0.99}            & \textbf{0.95}   \\ \hline
2023 Cohort  & 0.83      & \multicolumn{1}{c|}{\textbf{0.89}}   & \textbf{0.75}   & \multicolumn{1}{c|}{0.88}            & \textbf{0.83}   \\ \hline
Combined     & 0.90      & \multicolumn{1}{c|}{\textbf{0.93}}   & \textbf{0.84}   & \multicolumn{1}{c|}{0.90}            & \textbf{0.90}   \\ \hline
\end{tabular}%
}
\caption{Performance metrics for each model by cohort. Clinician prediction was collected at assessment prior to model inference. Model performance can be compared directly to clinicians using accuracy. We found that the stochastic model achieved clinician level performance across cohorts while the deterministic model fell behind even though the ROC AUC performance was higher across cohorts}
\label{tab:deployed}
\end{table}
\subsection{Dataset drift and model performance}
We observed considerable drift in the label distribution between cohorts. To recall, the label distribution of our training set consisted of 92\% for negative classes and 8\% for positive classes. In contrast, the 2021 and 2023 cohorts exhibited label distributions of 20\%-80\% and 64\%-36\%, respectively. We used a one-way Chi-square test to compare the relative frequencies of mortality outcomes in each cohort against the training distribution. The results revealed that for the 2021 cohort, $p < 0.01$, and for the 2023 cohort, $p > 0.01$.

\begin{figure*}[tb]
     \centering
     \begin{subfigure}[b]{0.48\textwidth}
         \centering
         \includegraphics[width=\textwidth]{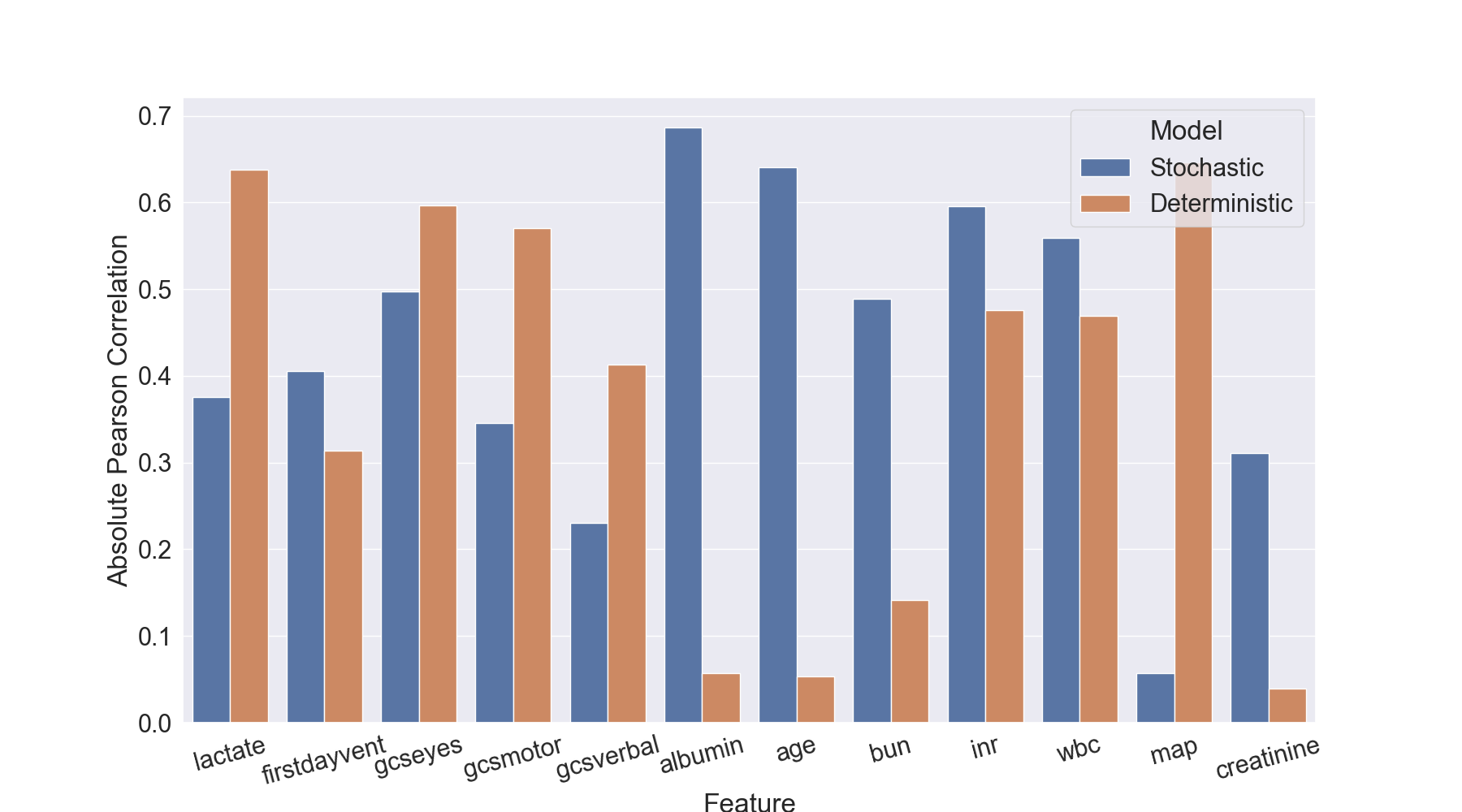}
         \caption{}
         \label{subfig:pearson}
     \end{subfigure}
     \hfill
     \begin{subfigure}[b]{0.48\textwidth}
         \centering
         \includegraphics[width=\textwidth]{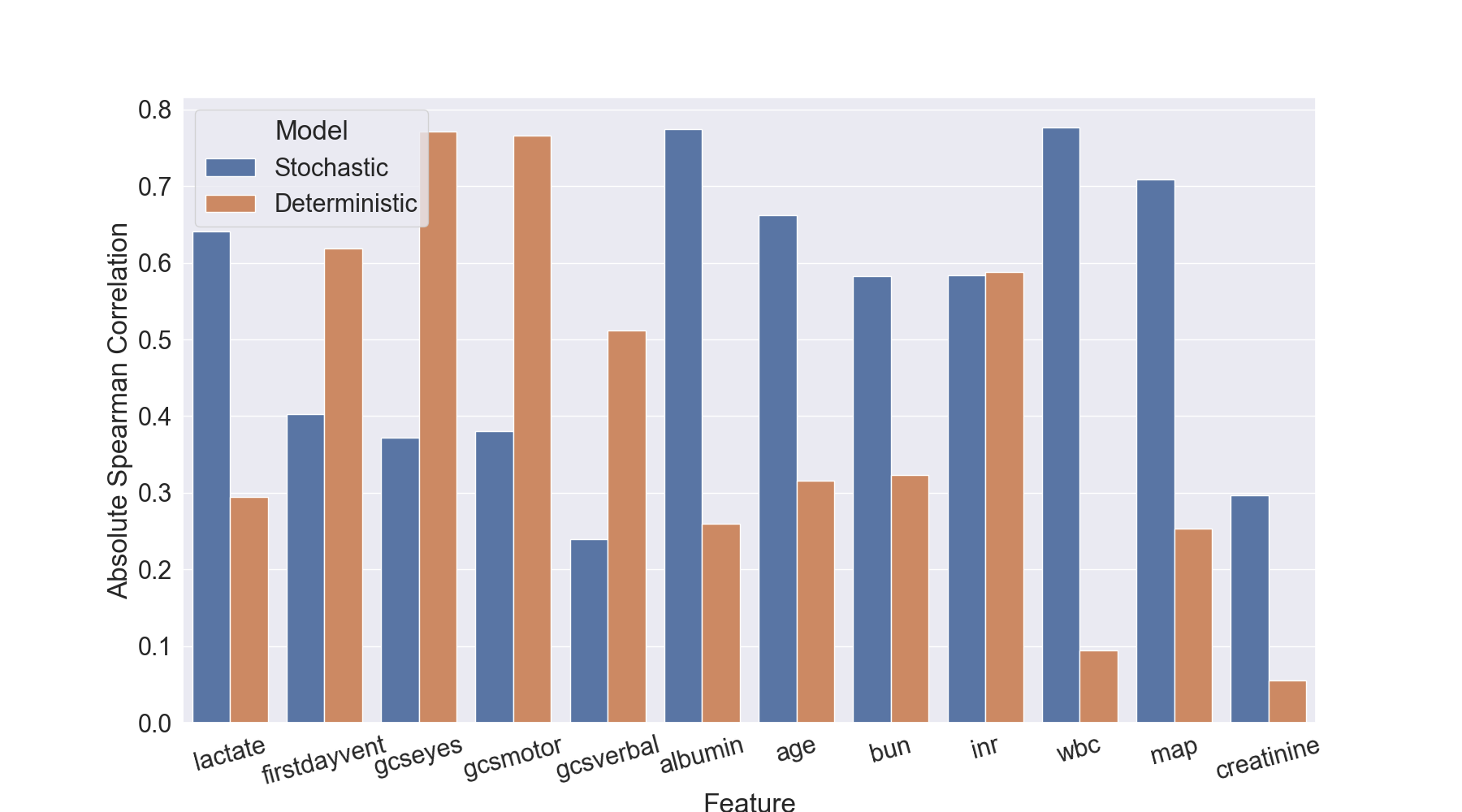}
         \caption{}
         \label{subfig:spearman}
     \end{subfigure}
    \caption{To validate influence function performance we adopted a validation scheme used by \cite{koh2017understanding} and computed the \textbf{\ref{subfig:pearson}:} absolute Pearson correlation and \textbf{\ref{subfig:spearman}:} absolute Spearman between pair-wise feature differences and pair-wise influence differences. 500 data pairs were randomly sampled from our deployment dataset. The explanations boast moderate to high correlation coefficients ($p<0.01$) for both models and metrics across most input features demonstrating the effectiveness of the method}
    \label{fig:exp}
\end{figure*}

To investigate the potential impact of dataset drift on performance, we utilized a 2-sample Kolmogorov-Smirnov test with Bonferroni correction to compare distributions. For all continuous features in our feature set, we assessed the distribution split by cohort relative to the training set distribution. Blood lactate and blood albumin were identified as significant ($p < 0.01$ for the 2021 cohort and $p > 0.01$ for the 2023 cohort), implying that the distribution of these variables significantly deviated from the training distribution for the 2021 cohort but not for the 2023 cohort. Kernel density estimates of these two features, separated by cohort are included in the supplementary material.

Turning our attention to model performance, we devoted substantial effort to optimizing the ROC AUC and LR+ during model selection, as these are conventional benchmarks for this problem due to the prevalent class imbalance. Since we also gather clinician predictions during data input, we can directly compare our models to clinicians using accuracy as a metric. Table \ref{tab:deployed} shows these results. We found that the stochastic model had achieved clinician level performance while the deterministic model fell behind across all cohorts even though the deterministic model had higher ROC AUC during model selection and across cohorts.

\subsection{Explainability}
To assess the explanations generated by both models, we performed a correlation analysis similar to that which was done in \cite{koh2017understanding}. We computed the correlation between pair-wise feature differences and pair-wise influence differences from our data collected from deployment by randomly sampling subjects and subject pairs without duplicate (N=500). Figure \ref{fig:exp} shows the absolute Pearson (Figure \ref{subfig:pearson}) and absolute Spearman (Figure \ref{subfig:spearman}) correlations. All correlations above 0.05 were significant ($p < 0.01$) with bonferonni correction. 

To better understand the kinds of explanations generated by each model, we examined the top three features by absolute magnitude for each subject and compared their frequency across both cohorts. The stochastic model displayed greater variability in its explanations, with nine features appearing in the top three positions for both cohorts. Figure \ref{fig:exp} shows the distribution of attribution scores for each feature by cohort. In contrast, the deterministic model only had six features in the top three positions. The stochastic model included lactate and albumin in it's top feature sets for both cohorts, which if you recall, were subject to significant distribution shift and therefore should be important.


In addition to importance scores, the influence function output is signed which can be used to evaluate class sentiment. To evaluate the overall sentiment associated with each feature, we applied the signum function to them. A highly positive/negative score would suggest that the model consistently attributes importance to one class, while a score closer to zero would indicate that the model is equally likely attribute importance to either class. Our findings revealed that the stochastic model had more values near zero compared to the deterministic model. The deterministic model exhibited mostly very positive or very negative values, indicating limited inter-subject variability. 


\subsection{Un(Certainty)}
To assess the effectiveness of our uncertainty metric, we conducted another 2-sample Kolmogorov-Smirnov test, comparing the ``Confidence'' metric reported by our stochastic model. This test revealed significant ($p < 0.01$) differences between the training set and the 2021 cohort, while no significant ($p > 0.01$) differences were observed between the training set and the 2023 cohort. Figure \ref{fig:sigma_dist} presents the kernel density estimates of the training distribution, first broken down by classifier correctness (Figure \ref{subfig:train_uncertainty}), and then by cohort (Figure \ref{subfig:cohort_uncertainty}). 

\section{Discussion}
\subsection{Robust models and the effect of COVID-19}
Given the substantial label shift in the data, we should have expected a sharp decline in performance. We attribute the robust performance of our models, in part, to the class-weighting/resampling approach we took during model selection. By resampling or weighting the loss function
(instead of changing the decision threshold), we blind our model to the inherent imbalance of the training dataset. Since, in both cohorts, the label shift actually makes the classes more balanced than the training set, we see an increase in performance.

The explanations provided by influence functions boast moderate to high correlation coefficients for both models across most input features demonstrating the effectiveness of the method. In particular, the explanations generated by the stochastic model displayed increased variability over the deterministic model, in both sentiment and magnitude, that may better capture the unique characteristics of different patients, potentially leading to more accurate and personalized predictions. This difference in behavior is consistent with literature on evaluating explanations produced by BNNs \cite{nielsen2023evalattai}. Future research should continue to explore and develop AI/ML models that can offer more detailed and individualized insights, ultimately enhancing their utility and applicability in real-world clinical scenarios.

Figure \ref{fig:sigma_dist} offers strong evidence supporting the robustness of BNNs. Our statistical results indicate that the ``Confidence'' metric may be used to detect dataset shift in small populations. The ability to effectively quantify uncertainty is crucial for AI/ML models in healthcare, as it allows clinicians to better understand the reliability of a given prediction and make more informed decisions based on the model's outputs. The visualizations in Figure \ref{fig:sigma_dist} demonstrate the utility of our uncertainty metric in identifying differences between various cohorts and the training set, providing valuable insights into the performance of our stochastic model across different conditions. As such, future research should continue to explore and refine methods for estimating uncertainty in AI/ML models, ensuring that these tools can provide healthcare professionals with accurate, reliable, and interpretable information to support their decision-making processes.

These findings underscore the importance of considering the deployment environment when developing AI/ML models. By utilizing stochastic models like BNNs that enhance the robustness of AI models, researchers can develop AI solutions that maintain their performance and reliability even in the face of changing conditions. This ultimately improves their utility and applicability in real-world clinical settings.

\subsection{Metric chasing}
Even though our models performed well, this was not the main focus of the effort. The prevailing trend in the majority of AI/ML research for healthcare involves developing models to predict various clinical endpoints to demonstrate incremental advancements in key performance metrics. However, only a few extend beyond these improvements to address other essential aspects of AI in healthcare. We firmly believe that future efforts should prioritize implementation science over merely enhancing the performance of predictive models.

Implementation science plays a crucial role in ensuring the successful deployment, adoption, and sustainability of AI systems in real-world healthcare settings. By focusing on this aspect, researchers can better understand the factors that facilitate the integration of evidence-based interventions and strategies into routine clinical practice. This, in turn, can lead to the development of AI solutions that not only excel in their predictive capabilities but also effectively address the unique challenges and complexities of healthcare environments.

Moreover, emphasizing implementation science can promote interdisciplinary collaboration, user-centered design, and stakeholder engagement, all of which contribute to the practical utility and long-term success of AI solutions in healthcare. By adopting a more comprehensive approach that encompasses these aspects, researchers can help bridge the gap between theoretical advancements and tangible improvements in clinical practice and patient outcomes.

\subsection{Limitations}
The most significant limitation of our study was the lack of integration into clinicians' workflow. At the outset, we underestimated the time and effort required for ICU clinicians to interrupt their tasks and enter 12 values on their phones. This challenge was further intensified by the COVID-19 pandemic, as clinicians faced longer working hours and managed more critical patients. If Electronic Health Record (EHR) integration had been possible, we estimate that we could have collected a considerably larger volume of data, potentially an order of magnitude greater than our current dataset.

While obtaining more data could have strengthened our results, we believe that it would not have fundamentally altered our findings. Given that our model was trained on 200,000 instances, we anticipate that our models demonstrate satisfactory calibration. Furthermore, the statistical testing we conducted produced positive results that corroborated our hypotheses.

In light of these limitations, future research should prioritize seamless integration into clinical workflows to facilitate data collection and ensure that AI models are practical and beneficial for healthcare professionals. Additionally, researchers should continue to explore methods for improving model robustness and generalizability, ensuring that AI solutions remain effective across diverse patient populations and varying medical conditions.

\section{Conclusion}
In this study, we examined the performance, explainability, and robustness of AI models for predicting clinical endpoints in the context of the COVID-19 pandemic. To our knowledge, this is the first study of its kind to deploy interpretable and explainable models with uncertainty prediction in a real-world setting. Our findings demonstrate the effectiveness of using BNNs to improve robustness, explanations and to quantify prediction uncertainty. By addressing these challenges, we were able to develop AI/ML models that matched or outperformed clinician predictions even when confronted with substantial dataset drift. Future research should continue to refine these methods to ensure these tools provide accurate, reliable, and interpretable information for healthcare professionals.


\appendix
\section{Variational Density Propagation (VDP) Derivation}
\label{sec:vdp_dev}
To describe the VDP method by Dera \emph{et al.} \cite{dera2019extended}, let us use a 2-layer fully connected neural network as an example. In a traditional or deterministic neural network, Equation \ref{eq:linear} describes the forward pass of the model.
\begin{equation}
    \begin{split}
        z = Wx+b^{[1]},\\
        a = f(z),\\
        \hat{y} = g(Va+b^{[2]}),
    \end{split}
    \label{eq:linear}
\end{equation}
\noindent where $W \in \mathbb{R}^{j\times k}$ is the weight matrix of layer 1, $x \in \mathbb{R}^{k \times 1}$ is the input vector, $b^{[1]} \in \mathbb{R}^{j\times 1}$ is the bias vector in layer-1, $z \in \mathbb{R}^{j \times 1}$ is the result of the linear operation in layer-1, $f$ is an arbitrary element-wise non-linear function, $g$ is an arbitrary non-linear activation function that does not operate element-wise, $a \in \mathbb{R}^{j \times 1}$ is the result after applying the non-linear activation function in layer 1, $V \in \mathbb{R}^{l \times j}$ is the weight matrix in layer 2, $b^{[2]} \in \mathbb{R}^{j \times 1}$ is the bias vector in layer 2, $\hat{y} \in \mathbb{R}^{l \times 1}$ is the predicted output and $k$ is the dimensionality of the input vector, $j$ is the number of nodes in layer 1 and $l$ is the number of classes to predict. 

To propagate the first two moments, several assumptions must be made. First, let us consider $w_m^\top  = m^{\text{th}}$ row of $W$, $m=1,2,...,j$. Then, $z_m = w_m^\top x+b_m^{[1]}$, $m=1,2,...,j$. Next, consider the following assumptions: the input vector $x$ is deterministic, $a \sim \mathcal{N}(\mu_a, \Sigma_a)$, $w_m \sim \mathcal{N}(\mu_{w_m}, \Sigma_{w_m})$, $m=1,2,...,j$, $b_m^{[1]} \sim \mathcal{N}(\mu_{b_m^{[1]}}, \sigma^2_{b_m^{[1]}})$, $m=1,2,...,j$, $v_n \sim \mathcal{N}(\mu_{v_n}, \Sigma_{v_n})$, $n=1,2,...,l$, $b_n^{[2]} \sim \mathcal{N}(\mu_{b_n}^{[2]}, \sigma^2_{b_n^{[2]}})$, $n=1,2,...,l$ and the weight vectors $w_m$, $a$, and bias $b^{[1]}$ and $b^{[2]}$ are uncorrelated to each other for $m=1,2,...,j$. The elements of $\mu_z$ and $\sigma^2_z$ are defined in Equations \ref{eq:muz} and \ref{eq:sigz}.

\begin{equation}
    \begin{split}
        \mu_{z_m} &= \mathbb{E}[w_m^\top x+b_m^{[1]}]\\
        &= \mathbb{E}[w_m^\top ]x + \mathbb{E}[b_m^{[1]}]\\
        &= \mu_{w_m}^\top  x + \mu_{b_m^{[1]}}
    \end{split}
    \label{eq:muz}
\end{equation}

\begin{equation}
    \begin{split}
        \sigma_{z_m}^2 &= \textrm{Var}[w_m^\top x+b_m^{[1]}]\\
        &= \textrm{Var}[w_m^\top x_m]+\textrm{Var}[b_m^{[1]}]\\
        &= x_m^\top \textrm{Var}[w_m]x_m+\textrm{Var}[b_m^{[1]}]\\
        &= x_m^\top \Sigma^2_{w_m}x_m+\sigma_{b_m^{[1]}}^2\\
    \end{split}
    \label{eq:sigz}
\end{equation}

\noindent Since we have assumed the weight vectors and elements of the bias $b$ to be uncorrelated, $\Sigma_{w_pw_q} = 0$ and $\sigma_{b_pb_q}=0$ for $p\neq q$, where $p, q = 1,2,...,j$. Hence, the covariance is zero. To propagate the first two moments through an arbitrary element-wise non-linear function, we utilize the first-order Taylor series approximation shown in Equations \ref{eq:elemmu} and \ref{eq:elemsig}.

\begin{equation}
    \begin{split}
        a &= f(z)\\
        &=f(\mu_z)+f'(\mu_z)(z-\mu_z)+ ...\\
        &\approx f(\mu_z)+f'(\mu_z)(z-\mu_z)\\
        \mathbb{E}[a] &= \mu_a \approx f(\mu_z)\\
        \mu_a &= f(\mu_z)
    \end{split}
    \label{eq:elemmu}
\end{equation}

\begin{equation}
    \Sigma_{a_pa_q} \approx \begin{cases}
    \sigma_{z_p}^2f'(\mu_z)^2, & p = q,\\
    \sigma_{z_pz_q}f'(\mu_{z_p})f'(\mu_{z_q}), & p \neq q.
    \end{cases}
    \label{eq:elemsig}
\end{equation}

For the second layer of the network, we can no longer consider the incoming vector, $\mu_a$, to be deterministic. Additionally we now need to propagate the incoming variance, $\sigma_a^2$. The mean and covariance propagated through the second fully-connected layer are given in Equations \ref{eq:muytilde} and \ref{eq:sigytilde}.

\begin{equation}
    \begin{split}
        \mu_{\tilde{y}} &= \mathbb{E}[v_n^\top  a+ b_n^{[2]}]\\
        &= \mathbb{E}[v_n^\top ]\mathbb{E}[a]+ \mathbb{E}[b_n^{[2]}]\\
        &=\mu_{v_n}^\top \mu_a+\mu_{b_n^{[2]}}
    \end{split}
    \label{eq:muytilde}
\end{equation}

\begin{equation}
    \Sigma_{\tilde{y}_p \Tilde{y}_q} =
        \begin{cases}
        \!\begin{aligned}
            &\textrm{Tr}(\Sigma_v\Sigma_a)+\mu_v^\top \Sigma_a\mu_v\\
            &+\mu_a^\top \Sigma_v\mu_a+\Sigma_{b_n^{[2]}}   
        \end{aligned}
        , & p=q,\\
        \mu_{v_p}^\top \Sigma_a \mu_{v_q}, & p \neq q.
        \end{cases}
    \label{eq:sigytilde}
\end{equation}

Next, we must use a non-linear activation function such as softmax to obtain the predictions of our model. Since $g$ does not operate element-wise on our mean and variance, we use a slightly different Taylor-series to obtain Equations \ref{eq:muyhat} and \ref{eq:sigyhat} \cite{simon2006optimal}.

\begin{equation}
    \mu_{\hat{y}} \approx g(\mu_{\tilde{y}})
    \label{eq:muyhat}
\end{equation}

\begin{equation}
    \Sigma_{\hat{y}} \approx J_g \Sigma_{\tilde{y}} J_g^\top ,
    \label{eq:sigyhat}
\end{equation}

\noindent where $J_g$ is the Jacobian matrix of the softmax function $g$ with respect to $\tilde{y}$ and calculated at $\mu_{\tilde{y}}$. 

Finally, we use the Evidence Lower Bound (ELBO) function, $\mathcal{L}(\phi, D)$, which consists of two parts: the expected log-likelihood of the training data given the weights, and a regularization term, $D=\{x^{(i)}, y^{(i)}\}_{i=1}^N$ given by Equation \ref{eq:elbo}.
\begin{equation}
    \mathcal{L}(\phi,D) = \mathbb{E}_{q(\phi)}[\log p(D|\phi)] - \textrm{KL}[q(\phi)|p(\phi)],
    \label{eq:elbo}
\end{equation}

\noindent where $\phi$ represents the weights $W$, $V$, and biases $b^{[1]}$, $b^{[2]}$. The expected log-likelihood is given in Equation \ref{eq:loglike} and the KL term is given in Equation \ref{eq:kl}.

\begin{equation}
\begin{split}
    &E_{q(\phi)}[\log p(D|\phi)] \approx \frac{1}{M} \sum_{m=1}^M \log p(D|\phi)\\
    &\approx -\frac{Nl}{2}\log(2\pi) - \frac{1}{M}\sum_{m=1}^M\biggl[\frac{N}{2}\log(|\Sigma_{\hat{y}}|)\\
    &\quad + \frac{1}{2}\sum_{i=1}^N(y^{(i)}-\mu_{\hat{y}}^{(m)})^\top (\Sigma_{\hat{y}}^{(m)})^{-1}(y^{(i)}-\mu_{\hat{y}}^{(m)})\biggr]
    \end{split}
    \label{eq:loglike}
\end{equation}

\noindent where, $y^{(i)}$ is the true label of the $i^{\textrm{th}}$ data point, $N$ is the number of data points and $M$ is the number of Monte Carlo samples needed to approximate the expectation by summation. 

\begin{equation}
    \textrm{KL}[q(\phi)|p(\phi)] = -\frac{1}{2}\sum_{n=1}^l(j\log \sigma^2_{v_n} - ||\mu_{v_n}||^2_F - j\sigma_{v_n}^2))
    \label{eq:kl}
\end{equation}

\section{Additional Model Selection Metrics}
\label{sec:metrics}
Table \ref{tab:training_res_appendix} shows the results of a 10-fold cross validation using the optimal parameters found for each model. The metric typically used to compare mortality prediction models is the area under the Receiver Operating Characteristic Area Under the Curve (ROC AUC). Both of our models achieve a significantly higher ROC AUC than the currently available and widely used by ICU Clinicians (APACHE \cite{zimmerman2006acute}, SAPS \cite{le1993new}), as well as a recently published Neural Network model \cite{epifano2020towards} ($p< 0.01$ for all pair-wise t-tests). An initially troublesome result was the low precision obtained by both models. Due to the loss function weighting/majority undersampling, both models predict the positive class (right or wrong) about 40\% of the time. This results in the model correctly classifying the positive class about 76\% of the time. Using Bayes rule, we can estimate a theoretical positive predictive value of 0.15, given the positive class only occurs 8\% of the time in the dataset. Therefore, our models are performing as expected given this imbalance.

\begin{table*}[ht]
\centering
\caption{10-fold cross validated training results for the traditional (deterministic) and Bayesian \cite{dera2019extended} (stochastic) neural nets. The mean score across the 10 runs is shown with the standard error. Parameter sweeps were run on both the eICU dataset \cite{pollard2018eicu} only as well as the union of the eICU \cite{pollard2018eicu} and MIMIC-III \cite{johnson2016mimic} datasets. The goal of the parameter sweep was to optimize positive likelihood ratio (LR+).}
\label{tab:training_res_appendix}
\resizebox{\textwidth}{!}{%
\begin{tabular}{|c|c|c|c|c|c|c|}
\hline
\textbf{Deterministic Network} & \textbf{Precision$\uparrow$} & \textbf{Sensitivity$\uparrow$} & \textbf{Specificity$\uparrow$} & \textbf{ROC AUC$\uparrow$} & \textbf{PRC AUC$\uparrow$} & \textbf{Balanced Accuracy$\uparrow$} \\ \hline
eICU Only                      & 0.21$\pm$(0.01)             & 0.79$\pm$(0.02)               & 0.82$\pm$(0.02)               & 0.89$\pm$(0.00)           & 0.43$\pm$(0.01)           & 0.81$\pm$(0.00)                                    \\ \hline
eICU + MIMIC-III               & 0.19$\pm$(0.00)             & 0.80$\pm$(0.00)               & 0.77$\pm$(0.00)               & 0.87$\pm$(0.00)           & 0.39$\pm$(0.00)           & 0.79$\pm$(0.00)                                    \\ \hline
\textbf{Stochastic Network}    & \textbf{Precision$\uparrow$} & \textbf{Sensitivity$\uparrow$} & \textbf{Specificity$\uparrow$} & \textbf{ROC AUC$\uparrow$} & \textbf{PRC AUC$\uparrow$} & \textbf{Balanced Accuracy$\uparrow$}                \\ \hline
eICU Only                      & 0.20$\pm$(0.01)               & 0.79$\pm$(0.02)                & 0.81$\pm$(0.02)                & 0.88$\pm$(0.00)            & 0.37$\pm$(0.02)            & 0.80$\pm$(0.00)                                     \\ \hline
eICU + MIMIC-III               & 0.19$\pm$(0.00)              & 0.80$\pm$(0.01)                & 0.79$\pm$(0.00)                & 0.87$\pm$(0.00)            & 0.35$\pm$(0.02)            & 0.79$\pm$(0.00)                                     \\ \hline
\end{tabular}%
}
\end{table*}

\subsection*{Variable Drift by Cohort}
\begin{figure*}[ht]
     \centering
     \begin{subfigure}[b]{0.48\textwidth}
         \centering
         \includegraphics[width=\textwidth]{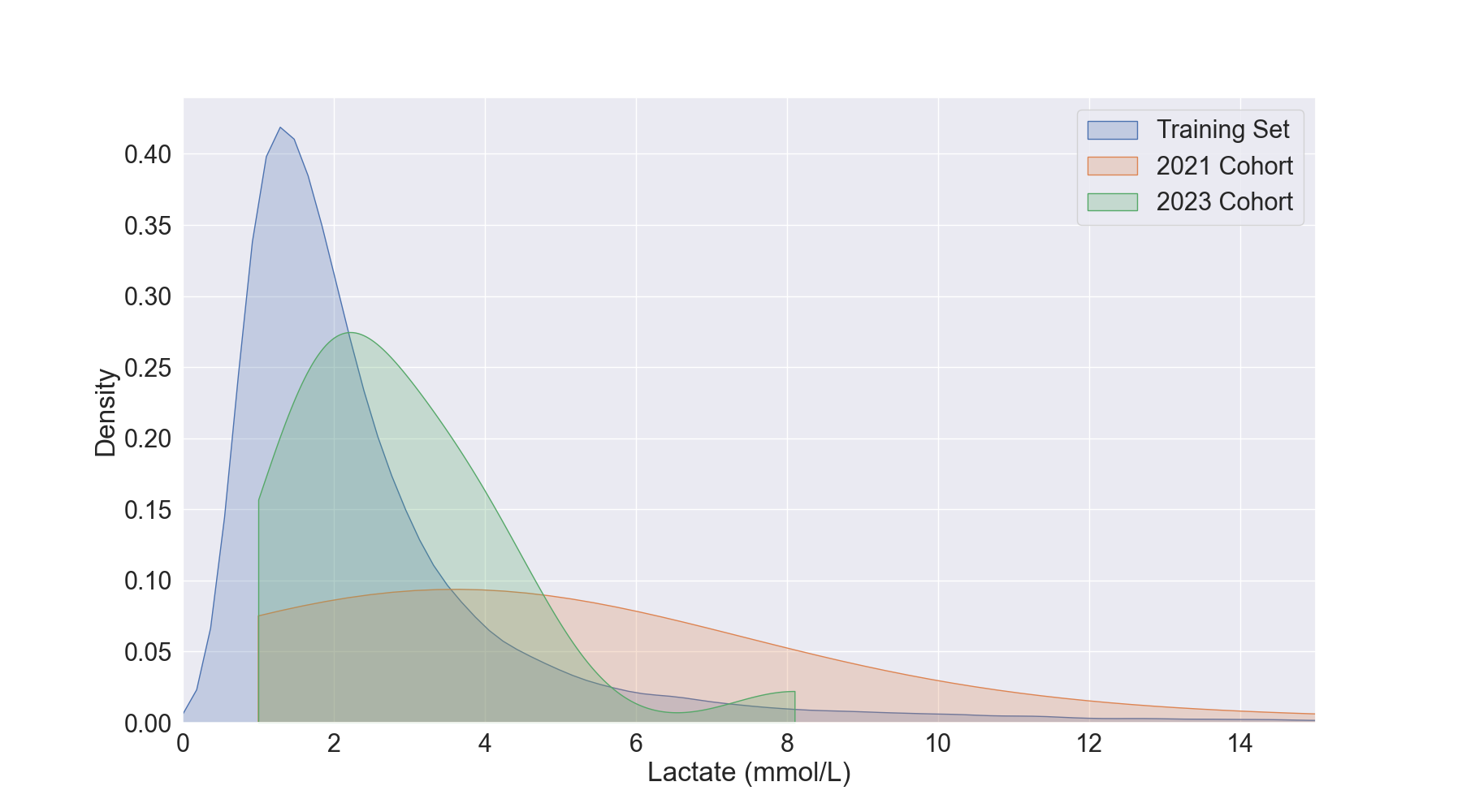}
         \caption{}
         \label{subfig:lactate}
     \end{subfigure}
     \hfill
     \begin{subfigure}[b]{0.48\textwidth}
         \centering
         \includegraphics[width=\textwidth]{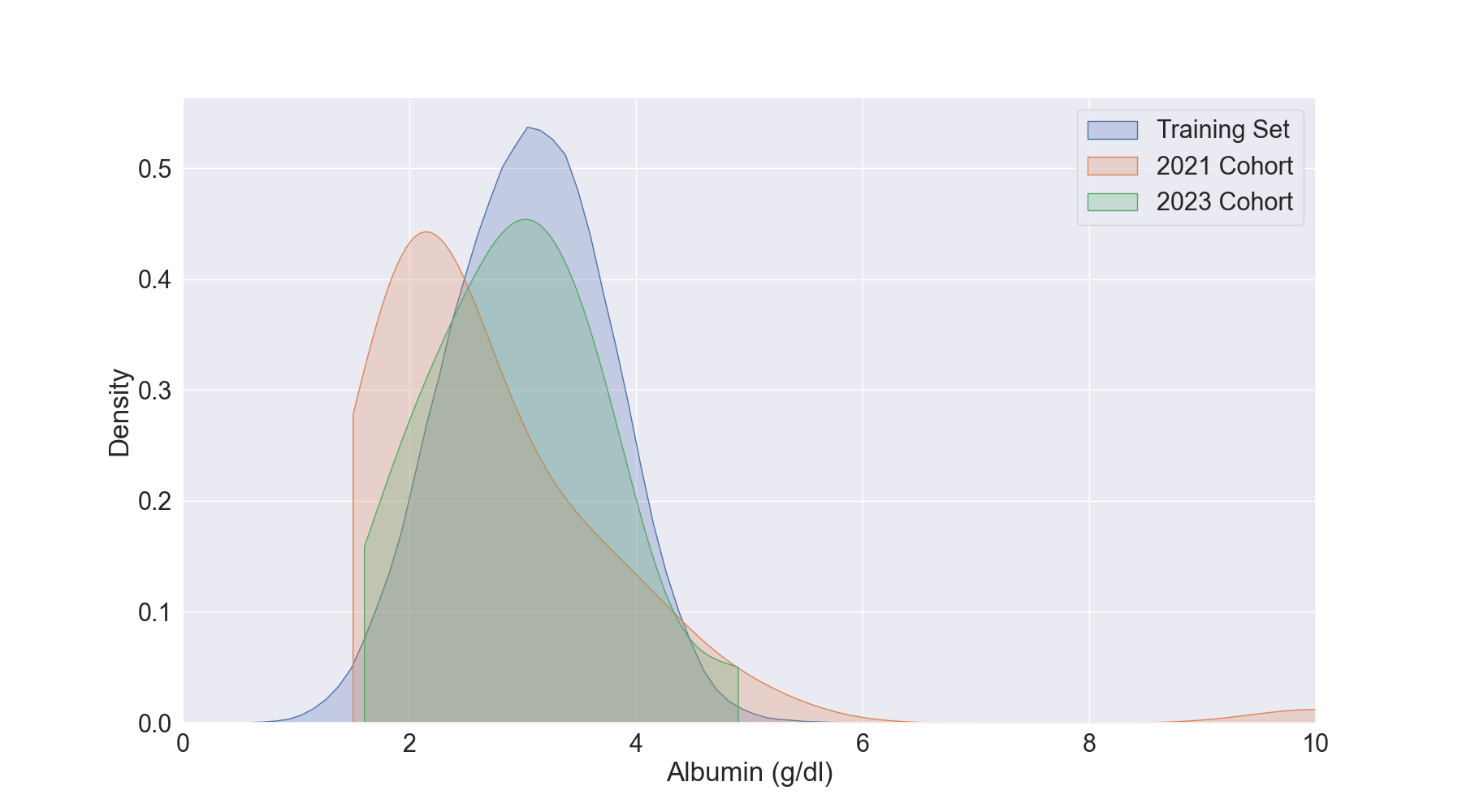}
         \caption{}
         \label{subfig:albumin}
     \end{subfigure}
    \caption{Kernel density estimates of two features that showed significant difference in distribution compared to the training set in 2021 but not in 2023 as a result of COVID-19 dataset drift ($p<0.01$ against the 2021 cohort and $p>0.01$ against the 2023 cohort using a 2-sample Kolmogorov–Smirnov test). \textbf{Left:} Lactate \textbf{Right:} Albumin}
    \label{fig:feats}
\end{figure*}

Figure \ref{fig:feats} shows kernel density estimate by cohort for the features that were determined by a 2-sample Kolmogorov–Smirnov test to be significantly different from the training set by the confidence metric from the stochastic model.


\subsubsection*{Acknowledgments}
\noindent Jacob R. Epifano is supported by US Department of Education GAANN award P200A180055. Ghulam Rasool was partly supported by NSF OAC-2008690.

 \bibliographystyle{unsrt} 
 \bibliography{references}

\begin{thebibliography}{10}

\bibitem{mesko2020short}
Bertalan Mesk{\'o} and Marton G{\"o}r{\"o}g.
\newblock A short guide for medical professionals in the era of artificial intelligence.
\newblock {\em npj Digital Medicine}, 3(1):1--8, 2020.

\bibitem{vayena2018machine}
Effy Vayena, Alessandro Blasimme, and I~Glenn Cohen.
\newblock Machine learning in medicine: addressing ethical challenges.
\newblock {\em PLoS medicine}, 15(11):e1002689, 2018.

\bibitem{smith2013ability}
Gary~B Smith, David~R Prytherch, Paul Meredith, Paul~E Schmidt, and Peter~I Featherstone.
\newblock The ability of the national early warning score (news) to discriminate patients at risk of early cardiac arrest, unanticipated intensive care unit admission, and death.
\newblock {\em Resuscitation}, 84(4):465--470, 2013.

\bibitem{masino2019machine}
Aaron~J Masino, Mary~Catherine Harris, Daniel Forsyth, Svetlana Ostapenko, Lakshmi Srinivasan, Christopher~P Bonafide, Fran Balamuth, Melissa Schmatz, and Robert~W Grundmeier.
\newblock Machine learning models for early sepsis recognition in the neonatal intensive care unit using readily available electronic health record data.
\newblock {\em PloS one}, 14(2):e0212665, 2019.

\bibitem{elish2018stakes}
Madeleine~Clare Elish.
\newblock The stakes of uncertainty: developing and integrating machine learning in clinical care.
\newblock In {\em Ethnographic Praxis in Industry Conference Proceedings}, volume 2018, pages 364--380. Wiley Online Library, 2018.

\bibitem{bedoya2019minimal}
Armando~D Bedoya, Meredith~E Clement, Matthew Phelan, Rebecca~C Steorts, Cara O’Brien, and Benjamin~A Goldstein.
\newblock Minimal impact of implemented early warning score and best practice alert for patient deterioration.
\newblock {\em Critical care medicine}, 47(1):49, 2019.

\bibitem{guidi2015clinician}
Jessica~L Guidi, Katherine Clark, Mark~T Upton, Hilary Faust, Craig~A Umscheid, Meghan~B Lane-Fall, Mark~E Mikkelsen, William~D Schweickert, Christine~A Vanzandbergen, Joel Betesh, et~al.
\newblock Clinician perception of the effectiveness of an automated early warning and response system for sepsis in an academic medical center.
\newblock {\em Annals of the American Thoracic Society}, 12(10):1514--1519, 2015.

\bibitem{dewan2020performance}
Maya Dewan, Naveen Muthu, Eric Shelov, Christopher~P Bonafide, Patrick Brady, Daniela Davis, Eric~S Kirkendall, Dana Niles, Robert~M Sutton, Danielle Traynor, et~al.
\newblock Performance of a clinical decision support tool to identify picu patients at high-risk for clinical deterioration.
\newblock {\em Pediatric critical care medicine: a journal of the Society of Critical Care Medicine and the World Federation of Pediatric Intensive and Critical Care Societies}, 21(2):129, 2020.

\bibitem{ginestra2019clinician}
Jennifer~C Ginestra, Heather~M Giannini, William~D Schweickert, Laurie Meadows, Michael~J Lynch, Kimberly Pavan, Corey~J Chivers, Michael Draugelis, Patrick~J Donnelly, Barry~D Fuchs, et~al.
\newblock Clinician perception of a machine learning-based early warning system designed to predict severe sepsis and septic shock.
\newblock {\em Critical care medicine}, 47(11):1477, 2019.

\bibitem{tonekaboni2019clinicians}
Sana Tonekaboni, Shalmali Joshi, Melissa~D McCradden, and Anna Goldenberg.
\newblock What clinicians want: contextualizing explainable machine learning for clinical end use.
\newblock In {\em Machine learning for healthcare conference}, pages 359--380. PMLR, 2019.

\bibitem{jimenez2020drug}
Jos{\'e} Jim{\'e}nez-Luna, Francesca Grisoni, and Gisbert Schneider.
\newblock Drug discovery with explainable artificial intelligence.
\newblock {\em Nature Machine Intelligence}, 2(10):573--584, 2020.

\bibitem{cutillo2020machine}
Christine~M Cutillo, Karlie~R Sharma, Luca Foschini, Shinjini Kundu, Maxine Mackintosh, Kenneth~D Mandl, and MI~in~Healthcare Workshop Working Group Beck Tyler 1 Collier Elaine 1 Colvis Christine 1 Gersing Kenneth 1 Gordon Valery 1 Jensen Roxanne 8 Shabestari Behrouz 9 Southall Noel~1.
\newblock Machine intelligence in healthcare—perspectives on trustworthiness, explainability, usability, and transparency.
\newblock {\em NPJ digital medicine}, 3(1):47, 2020.

\bibitem{combi2022manifesto}
Carlo Combi, Beatrice Amico, Riccardo Bellazzi, Andreas Holzinger, Jason~H Moore, Marinka Zitnik, and John~H Holmes.
\newblock A manifesto on explainability for artificial intelligence in medicine.
\newblock {\em Artificial Intelligence in Medicine}, 133:102423, 2022.

\bibitem{london2019artificial}
Alex~John London.
\newblock Artificial intelligence and black-box medical decisions: accuracy versus explainability.
\newblock {\em Hastings Center Report}, 49(1):15--21, 2019.

\bibitem{nielsen2023evalattai}
Ian~E Nielsen, Ravi~P Ramachandran, Nidhal Bouaynaya, Hassan~M Fathallah-Shaykh, and Ghulam Rasool.
\newblock Evalattai: A holistic approach to evaluating attribution maps in robust and non-robust models.
\newblock {\em IEEE Access}, 2023.

\bibitem{rudin2019stop}
Cynthia Rudin.
\newblock Stop explaining black box machine learning models for high stakes decisions and use interpretable models instead.
\newblock {\em Nature Machine Intelligence}, 1(5):206--215, 2019.

\bibitem{nguyen2015deep}
Anh Nguyen, Jason Yosinski, and Jeff Clune.
\newblock Deep neural networks are easily fooled: High confidence predictions for unrecognizable images.
\newblock In {\em Proceedings of the IEEE conference on computer vision and pattern recognition}, pages 427--436, 2015.

\bibitem{guo2017calibration}
Chuan Guo, Geoff Pleiss, Yu~Sun, and Kilian~Q Weinberger.
\newblock On calibration of modern neural networks.
\newblock In {\em International conference on machine learning}, pages 1321--1330. PMLR, 2017.

\bibitem{ahmed2022failure}
Sabeen Ahmed, Dimah Dera, Saud~Ul Hassan, Nidhal Bouaynaya, and Ghulam Rasool.
\newblock Failure detection in deep neural networks for medical imaging.
\newblock {\em Frontiers in Medical Technology}, 4, 2022.

\bibitem{umscheid2015development}
Craig~A Umscheid, Joel Betesh, Christine VanZandbergen, Asaf Hanish, Gordon Tait, Mark~E Mikkelsen, Benjamin French, and Barry~D Fuchs.
\newblock Development, implementation, and impact of an automated early warning and response system for sepsis.
\newblock {\em Journal of hospital medicine}, 10(1):26--31, 2015.

\bibitem{embi2012evaluating}
Peter~J Embi and Anthony~C Leonard.
\newblock Evaluating alert fatigue over time to ehr-based clinical trial alerts: findings from a randomized controlled study.
\newblock {\em Journal of the American Medical Informatics Association}, 19(e1):e145--e148, 2012.

\bibitem{kelly2019key}
Christopher~J Kelly, Alan Karthikesalingam, Mustafa Suleyman, Greg Corrado, and Dominic King.
\newblock Key challenges for delivering clinical impact with artificial intelligence.
\newblock {\em BMC medicine}, 17(1):1--9, 2019.

\bibitem{nestor2018rethinking}
Bret Nestor, Matthew McDermott, Geeticka Chauhan, Tristan Naumann, Michael~C Hughes, Anna Goldenberg, and Marzyeh Ghassemi.
\newblock Rethinking clinical prediction: why machine learning must consider year of care and feature aggregation.
\newblock {\em arXiv preprint arXiv:1811.12583}, 2018.

\bibitem{chen2018my}
Irene Chen, Fredrik~D Johansson, and David Sontag.
\newblock Why is my classifier discriminatory?
\newblock {\em Advances in neural information processing systems}, 31, 2018.

\bibitem{koh2017understanding}
Pang~Wei Koh and Percy Liang.
\newblock Understanding black-box predictions via influence functions.
\newblock In {\em International conference on machine learning}, pages 1885--1894. PMLR, 2017.

\bibitem{epifano2020towards}
Jacob~R Epifano, Ravi~P Ramachandran, Sharad Patel, and Ghulam Rasool.
\newblock Towards an explainable mortality prediction model.
\newblock In {\em 2020 IEEE 30th International Workshop on Machine Learning for Signal Processing (MLSP)}, pages 1--6. IEEE, 2020.

\bibitem{lundberg2017unified}
Scott~M Lundberg and Su-In Lee.
\newblock A unified approach to interpreting model predictions.
\newblock {\em Advances in neural information processing systems}, 30, 2017.

\bibitem{rodriguez2019interpretation}
Raquel Rodr{\'\i}guez-P{\'e}rez and J{\"u}rgen Bajorath.
\newblock Interpretation of compound activity predictions from complex machine learning models using local approximations and shapley values.
\newblock {\em Journal of medicinal chemistry}, 63(16):8761--8777, 2019.

\bibitem{safaei2022catboost}
Nima Safaei, Babak Safaei, Seyedhouman Seyedekrami, Mojtaba Talafidaryani, Arezoo Masoud, Shaodong Wang, Qing Li, and Mahdi Moqri.
\newblock E-catboost: An efficient machine learning framework for predicting icu mortality using the eicu collaborative research database.
\newblock {\em Plos one}, 17(5):e0262895, 2022.

\bibitem{hullermeier2021aleatoric}
Eyke H{\"u}llermeier and Willem Waegeman.
\newblock Aleatoric and epistemic uncertainty in machine learning: An introduction to concepts and methods.
\newblock {\em Machine Learning}, 110(3):457--506, 2021.

\bibitem{lakshminarayanan2017simple}
Balaji Lakshminarayanan, Alexander Pritzel, and Charles Blundell.
\newblock Simple and scalable predictive uncertainty estimation using deep ensembles.
\newblock {\em Advances in neural information processing systems}, 30, 2017.

\bibitem{blundell2015weight}
Charles Blundell, Julien Cornebise, Koray Kavukcuoglu, and Daan Wierstra.
\newblock Weight uncertainty in neural network.
\newblock In {\em International conference on machine learning}, pages 1613--1622. PMLR, 2015.

\bibitem{dera2019extended}
Dimah Dera, Ghulam Rasool, and Nidhal Bouaynaya.
\newblock Extended variational inference for propagating uncertainty in convolutional neural networks.
\newblock In {\em 2019 IEEE 29th International Workshop on Machine Learning for Signal Processing (MLSP)}, pages 1--6. IEEE, 2019.

\bibitem{izmailov2018averaging}
Pavel Izmailov, Dmitrii Podoprikhin, Timur Garipov, Dmitry Vetrov, and Andrew~Gordon Wilson.
\newblock Averaging weights leads to wider optima and better generalization.
\newblock {\em arXiv preprint arXiv:1803.05407}, 2018.

\bibitem{kompa2021second}
Benjamin Kompa, Jasper Snoek, and Andrew~L Beam.
\newblock Second opinion needed: communicating uncertainty in medical machine learning.
\newblock {\em NPJ Digital Medicine}, 4(1):4, 2021.

\bibitem{auld2022trends}
Sara~C Auld, Kristin~RV Harrington, Max~W Adelman, Chad~J Robichaux, Elizabeth~C Overton, Mark Caridi-Scheible, Craig~M Coopersmith, and David~J Murphy.
\newblock Trends in icu mortality from coronavirus disease 2019: a tale of three surges.
\newblock {\em Critical Care Medicine}, 50(2):245, 2022.

\bibitem{yan2020interpretable}
Li~Yan, Hai-Tao Zhang, Jorge Goncalves, Yang Xiao, Maolin Wang, Yuqi Guo, Chuan Sun, Xiuchuan Tang, Liang Jing, Mingyang Zhang, et~al.
\newblock An interpretable mortality prediction model for covid-19 patients.
\newblock {\em Nature machine intelligence}, 2(5):283--288, 2020.

\bibitem{quanjel2021replication}
Marian~JR Quanjel, Thijs~C van Holten, Pieternel~C Gunst-van~der Vliet, Jette Wielaard, Bekir Karakaya, Maaike S{\"o}hne, Hazra~S Moeniralam, and Jan~C Grutters.
\newblock Replication of a mortality prediction model in dutch patients with covid-19.
\newblock {\em Nature Machine Intelligence}, 3(1):23--24, 2021.

\bibitem{davis2019nonparametric}
Sharon~E Davis, Robert~A Greevy~Jr, Christopher Fonnesbeck, Thomas~A Lasko, Colin~G Walsh, and Michael~E Matheny.
\newblock A nonparametric updating method to correct clinical prediction model drift.
\newblock {\em Journal of the American Medical Informatics Association}, 26(12):1448--1457, 2019.

\bibitem{johnson2016mimic}
Alistair~EW Johnson, Tom~J Pollard, Lu~Shen, Li-wei~H Lehman, Mengling Feng, Mohammad Ghassemi, Benjamin Moody, Peter Szolovits, Leo Anthony~Celi, and Roger~G Mark.
\newblock Mimic-iii, a freely accessible critical care database.
\newblock {\em Scientific data}, 3(1):1--9, 2016.

\bibitem{pollard2018eicu}
Tom~J Pollard, Alistair~EW Johnson, Jesse~D Raffa, Leo~A Celi, Roger~G Mark, and Omar Badawi.
\newblock The eicu collaborative research database, a freely available multi-center database for critical care research.
\newblock {\em Scientific data}, 5(1):1--13, 2018.

\bibitem{van2011mice}
Stef Van~Buuren and Karin Groothuis-Oudshoorn.
\newblock mice: Multivariate imputation by chained equations in r.
\newblock {\em Journal of statistical software}, 45:1--67, 2011.

\bibitem{wandb}
Lukas Biewald.
\newblock Experiment tracking with weights and biases, 2020.
\newblock Software available from wandb.com.

\bibitem{deeks2004diagnostic}
Jonathan~J Deeks and Douglas~G Altman.
\newblock Diagnostic tests 4: likelihood ratios.
\newblock {\em Bmj}, 329(7458):168--169, 2004.

\bibitem{chawla2002smote}
Nitesh~V Chawla, Kevin~W Bowyer, Lawrence~O Hall, and W~Philip Kegelmeyer.
\newblock Smote: synthetic minority over-sampling technique.
\newblock {\em Journal of artificial intelligence research}, 16:321--357, 2002.

\bibitem{JMLR:v18:16-365}
Guillaume Lema{{\^i}}tre, Fernando Nogueira, and Christos~K. Aridas.
\newblock Imbalanced-learn: A python toolbox to tackle the curse of imbalanced datasets in machine learning.
\newblock {\em Journal of Machine Learning Research}, 18(17):1--5, 2017.

\bibitem{dera2020bayes}
Dimah Dera, Ghulam Rasool, Nidhal~C Bouaynaya, Adam Eichen, Stephen Shanko, Jeff Cammerata, and Sanipa Arnold.
\newblock {Bayes-SAR net: Robust SAR image classification with uncertainty estimation using bayesian convolutional neural network}.
\newblock In {\em 2020 IEEE International Radar Conference (RADAR)}, pages 362--367. IEEE, 2020.

\bibitem{dera2021premium}
Dimah Dera, Nidhal Bouaynaya, Ghulam Rasool, R~Shterenberg, and Hassan Fathallah-Shaykh.
\newblock Premium-cnn: Propagating uncertainty towards robust convolutional neural networks.
\newblock {\em IEEE Transactions on Signal Processing}, 2021.

\bibitem{manceur2013maximum}
Ameur~M Manceur and Pierre Dutilleul.
\newblock Maximum likelihood estimation for the tensor normal distribution: Algorithm, minimum sample size, and empirical bias and dispersion.
\newblock {\em Journal of Computational and Applied Mathematics}, 239:37--49, 2013.

\bibitem{le1993new}
Jean-Roger Le~Gall, Stanley Lemeshow, and Fabienne Saulnier.
\newblock A new simplified acute physiology score (saps ii) based on a european/north american multicenter study.
\newblock {\em Jama}, 270(24):2957--2963, 1993.

\bibitem{zimmerman2006acute}
Jack~E Zimmerman, Andrew~A Kramer, Douglas~S McNair, and Fern~M Malila.
\newblock Acute physiology and chronic health evaluation (apache) iv: hospital mortality assessment for today’s critically ill patients.
\newblock {\em Critical care medicine}, 34(5):1297--1310, 2006.

\bibitem{epifanofeature}
Jacob~R. Epifano, Alison Silvestri, Alexander Yu, Ravi~P. Ramachandran, Aakash Tripathi, and Ghulam Rasool.
\newblock A comparison of feature selection techniques for first-day mortality prediction in the icu.
\newblock In {\em 2023 IEEE International Symposium on Circuits and Systems (ISCAS)}, pages 1--5, 2023.

\bibitem{basu2020influence}
Samyadeep Basu, Philip Pope, and Soheil Feizi.
\newblock Influence functions in deep learning are fragile.
\newblock {\em arXiv preprint arXiv:2006.14651}, 2020.

\bibitem{epifano2023revisiting}
Jacob~R Epifano, Ravi~P Ramachandran, Aaron~J Masino, and Ghulam Rasool.
\newblock Revisiting the fragility of influence functions.
\newblock {\em Neural Networks}, 162:581--588, 2023.

\bibitem{ribeiro2016should}
Marco~Tulio Ribeiro, Sameer Singh, and Carlos Guestrin.
\newblock " why should i trust you?" explaining the predictions of any classifier.
\newblock In {\em Proceedings of the 22nd ACM SIGKDD international conference on knowledge discovery and data mining}, pages 1135--1144, 2016.

\bibitem{slack2020fooling}
Dylan Slack, Sophie Hilgard, Emily Jia, Sameer Singh, and Himabindu Lakkaraju.
\newblock Fooling lime and shap: Adversarial attacks on post hoc explanation methods.
\newblock In {\em Proceedings of the AAAI/ACM Conference on AI, Ethics, and Society}, pages 180--186, 2020.

\bibitem{lakkaraju2020fool}
Himabindu Lakkaraju and Osbert Bastani.
\newblock " how do i fool you?" manipulating user trust via misleading black box explanations.
\newblock In {\em Proceedings of the AAAI/ACM Conference on AI, Ethics, and Society}, pages 79--85, 2020.

\bibitem{fernandez2020explaining}
Carlos Fern{\'a}ndez-Lor{\'\i}a, Foster Provost, and Xintian Han.
\newblock Explaining data-driven decisions made by ai systems: the counterfactual approach.
\newblock {\em arXiv preprint arXiv:2001.07417}, 2020.

\bibitem{simon2006optimal}
Dan Simon.
\newblock {\em Optimal state estimation: Kalman, H infinity, and nonlinear approaches}.
\newblock John Wiley \& Sons, 2006.

\end{thebibliography}

\end{document}